\documentclass[10pt,twocolumn,letterpaper]{article}

\usepackage{cvpr}              
\usepackage[accsupp]{axessibility}  

\usepackage{url}
\usepackage{booktabs}
\usepackage{xcolor}
\usepackage{colortbl}
\usepackage{graphicx}
\usepackage{mathtools, nccmath}
\usepackage{wrapfig}
\usepackage{xspace}

\definecolor{orange}{rgb}{1,0.5,0}
\definecolor{gr}{rgb}{0,0.65,0}
\definecolor{mygray}{gray}{0.95}
\definecolor{lgray}{gray}{0.9}

\def\x{{\mathbf{x}}}

\usepackage{amsmath,amsfonts,bm}
\usepackage[symbol]{footmisc}

\definecolor{cvprblue}{rgb}{0.21,0.49,0.74}
\usepackage[pagebackref,breaklinks,colorlinks,allcolors=cvprblue]{hyperref}

\def\paperID{*****} 
\def\confName{CVPR}
\def\confYear{2026}

\title{Video Panels for Long Video Understanding}

\author{Lars Doorenbos\textsuperscript{*1,2}, Federico Spurio\textsuperscript{*1,2}, Juergen Gall\textsuperscript{1,2}\\
\textsuperscript{1}University of Bonn \quad \textsuperscript{2}Lamarr Institute for Machine Learning and Artificial Intelligence\\
{\tt\small \{doorenbos,spurio,gall\}@iai.uni-bonn.de}
}
\begin{document}

\maketitle
\footnotetext[1]{Equal contribution.}

\begin{abstract}
Recent Video-Language Models (VLMs) achieve promising results on long-video understanding, but their performance still lags behind that achieved on tasks involving images or short videos. This has led to great interest in improving the long context modeling of VLMs by introducing novel modules and additional complexity. 
In this paper, we take a different approach: rather than fine-tuning VLMs with the limited data available, we attempt to maximize the performance of existing models.
To this end, we propose a novel visual prompting strategy specifically designed for long-video understanding. By combining multiple frames as panels into one image, we effectively trade off spatial details for temporal resolution.
Our approach is training-free, parameter-free, and model-agnostic, and can be seamlessly integrated into existing VLMs.
Extensive experiments on five established benchmarks across a wide range of model architectures, sizes, and context windows confirm the consistency of our approach. For the TimeScope (Long) dataset, which has the longest videos, the accuracy for video question answering is improved by up to 19.4\%. Overall, our method raises the bar for long video understanding models. The code is available at \url{https://fedespu.github.io/Video-Panels}.  
\end{abstract}    
\section{Introduction}

\begin{figure*}[tb]
    \centering
    \includegraphics[width=\linewidth]{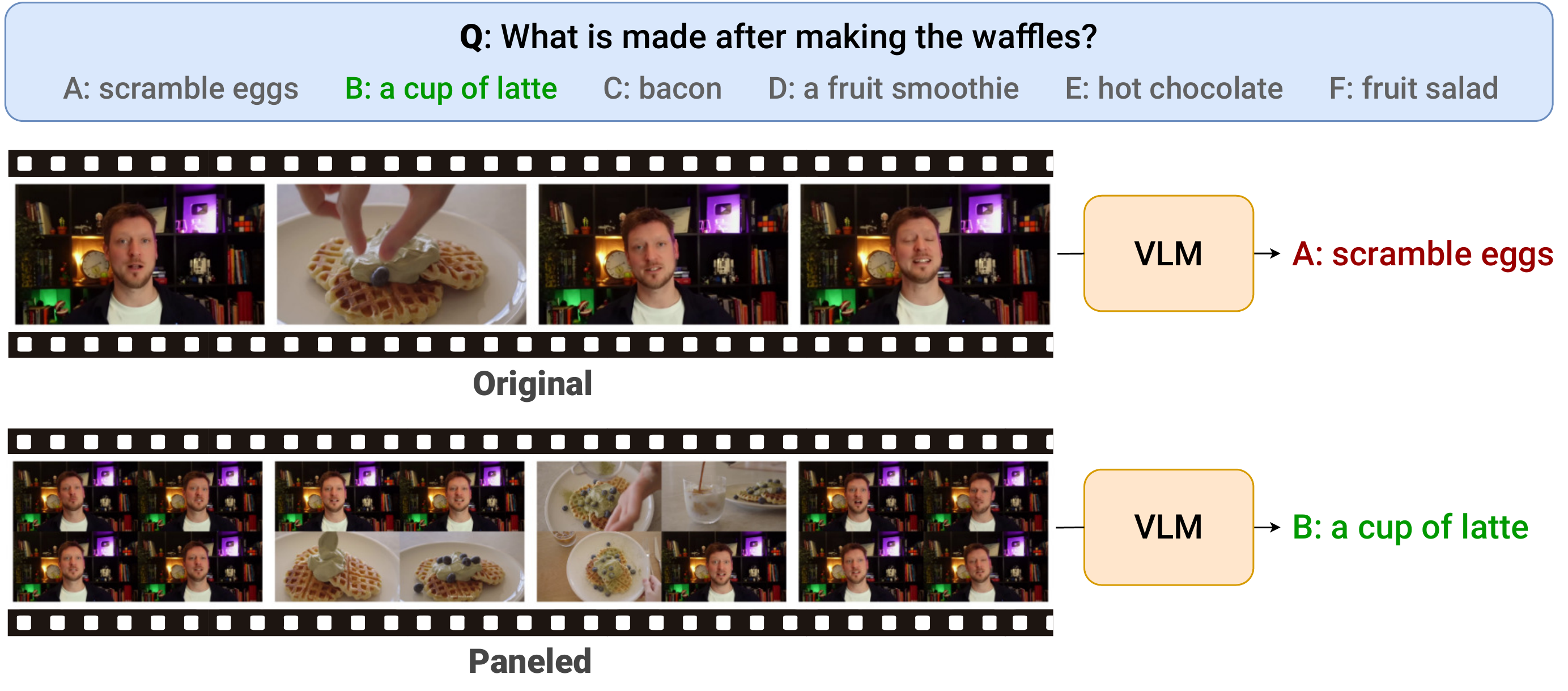}
    \caption{\textbf{Illustration of our approach.} We show the output of LLaVA-OneVision 7B on a sample from Video-MME. Without our proposed visual prompting, the model is unable to answer the question correctly. By paneling multiple frames into one, we increase the capacity for long video understanding and output the correct answer, without introducing any additional complexity. }
    \label{fig:method}
\end{figure*}

Connecting the reasoning capabilities of Large Language Models (LLMs) with other modalities has opened up the possibility to reason about and interact with multi-modal data~\citep{yin2024survey,Yi_2025_CVPR}. In particular Vision–Language Models, which are capable of reasoning over images or videos based on a textual query, now have impressive perceptive and cognitive capabilities. This has led to their application across a variety of imaging domains like medical imaging~\citep{hartsock2024vision} and industrial defect detection~\citep{mokhtar2025detect}, and progress is further driven by the publication of strong open-source models such as LLaVA-Video~\citep{zhangllava}, LLaVA-OneVision~\citep{li2024llava}, and Qwen2.5-VL~\citep{bai2025qwen2}. However, with image datasets and benchmarks far outnumbering those for videos, video understanding by Video–Language Models (VLMs) remains a challenge. 

While VLMs can still achieve high performance on tasks involving short video clips, their performance drastically decreases as video length increases. For example, Qwen-VL2.5 exhibits a notable drop in accuracy when processing videos longer than three minutes~\citep{timescope}. This decline is largely caused by their limited temporal resolution, stemming from relatively small context windows and memory constraints, which restrict the effective visual context that the model can process. 
Recent works thus try to improve the long context modeling of VLMs in a variety of ways, e.g., by heavily compressing the tokens used to represent input frames before modeling the temporal dynamics~\citep{bai2025qwen2,li2024videochat,zhang2025videollama}. Nevertheless, many of them struggle to consistently outperform the base models with limited context sizes~\citep{shu2025video,wang2025videollamb,zhang2024long}. 
We therefore address the question: \textit{Is there a simple yet efficient way to consistently improve long video understanding of existing VLMs?} 


In this paper, we thus take a different approach. Rather than fine-tuning VLMs for long-video understanding with the limited instruction data available, we focus on making the most out of the capabilities of existing models. To this end, we introduce the first visual prompting approach designed specifically for long video understanding. Specifically, we combine multiple frames of a video into one image, as panels in a comic as shown in Fig.~\ref{fig:method}. Our method leverages these multi-panel inputs to improve temporal reasoning: a sequence of such panels enables denser temporal coverage within the model’s constrained input window. This effectively increases temporal resolution at the cost of some spatial detail, leading to a better balance between them for long videos. Since our approach does not modify the underlying VLM architecture and is training-free and model-agnostic, it is a versatile approach that can be applied to any existing VLM. 

We evaluate our prompting approach on 5 video question answering benchmarks, namely VideoMME~\citep{fu2025video}, TimeScope~\citep{timescope}, MLVU~\citep{zhou2025mlvu}, 
MF2~\citep{zaranis2025movie}, and VNBench~\citep{zhao2024needle}, and apply our approach to 8 VLMs, namely Video-LLaVA~\citep{lin2023video}, VideoChat2-HD~\citep{li2024mvbench}, LLaVA-OV~\citep{li2024llava}, Qwen-2VL~\citep{wang2024qwen2}, Qwen-2.5VL~\citep{bai2025qwen2}, LLaVA-Video~\citep{zhangllava}, VideoLLaMA 3~\citep{zhang2024long}, and GPT-4~\cite{achiam2023gpt}, which use between 8 and 180 context frames. Our approach improves the base VLMs in most settings and, on average, across all datasets for all VLMs. For the TimeScope (Long) dataset, which contains the longest videos, the accuracy of VideoLLaMA 3, which is the strongest base model on this dataset, is improved by a large margin. The accuracy is increased by $+7.6$, which is an improvement by ${19.4\%}$.    
This shows that paneled images are a simple yet efficient approach to improve the long video understanding performance of VLMs. 

Beyond zero-shot inference, we also explore whether panels provide a more effective representation for long videos in general. By fine-tuning VLMs on their original training data represented as panels, we achieve additional performance gains on long-video benchmarks, demonstrating that video panels complement training-based improvements rather than replacing them.
Overall, the main contributions of this paper are as follows:
\begin{itemize}
    \item We present the first visual prompt engineering method for long-video understanding that requires no training or additional parameters, and can be seamlessly integrated into existing VLMs.
    \item We show through extensive experiments that our framework consistently improves results across a wide range of benchmarks, model types, and context sizes.
    \item We show that fine-tuning can increase the performance even further, without introducing new training data.
\end{itemize}

\section{Related Work}

\paragraph{Long video understanding.}
Scaling vision–language models (VLMs) to long videos is challenging due to the large number of visual tokens and the limited context windows of language backbones. A common strategy is to reduce the input size, either by modifying visual re-samplers to extract fewer tokens~\citep{li2023blip, li2024llama, cheng2024videollama} or by pruning and merging features through heuristic techniques~\citep{chen2024image, shang2024llava, jin2024chat, zhou2024streaming}. Other works extend the sequence capacity of the LLM, as in LongVA~\citep{zhang2024long}, who transfer the long-context pretraining of their language backbones to multi-modal settings.
Another line of work aims to preserve temporal coherence through memory mechanisms or summarization tokens~\cite{jin2024chat,song2024moviechat,wei2025visual}. For instance, VideoLLaMB~\citep{wang2025videollamb} introduces recurrent memory bridges with scene segmentation, Video-XL~\citep{shu2025video} condenses video intervals into summarization tokens within the Transformer, and SF-LLaVA~\cite{xu2024slowfast} uses both a slow and fast pathway to balance spatial detail with long-term context. 
Overall, this variety in strategies highlights the difficulty in finding a good balance between efficiency and information preservation. 

A key insight is that these methods operate at the token level. However, our ablation studies in Sec.~\ref{sec:why} show that pooling frames, rather than tokens, yields better performance. Moreover, previous approaches are model-specific, which limits their general applicability. 
Together, these limitations motivate us to explore a more general alternative in the form of visual prompting.

\paragraph{Visual prompting in VLMs.}
Visual prompting, also known as visual prompt engineering, is a strategy that modifies the visual input (images or videos) of VLMs to direct their attention, rather than altering the model itself. Prior studies show that even simple geometric cues, such as \eg, colored regions~\citep{yao2024cpt} or red circles~\citep{shtedritski2023does} for CLIP~\citep{radford2021learning}, or bounding boxes~\citep{dang2023instructdet, duan2024cityllava, ma2024groma}, can effectively guide focus in image understanding tasks. In multi-modal settings, visual prompting has been successfully applied to improve alignment and reasoning in MLLMs~\citep{cai2024vip, wu2024visual}. Extending this idea to videos, \cite{wu2025number} introduces numerical tags as visual prompts within video frames, \cite{du2025motionsight} improves fine-grained motion recognition with motion blur and spotlighting, and~\cite{wang2025visual} improves emotion recognition with an ensemble of visual prompting techniques. Our work follows a related intuition: we project temporal information into the spatial domain by concatenating subsequent frames into a single visual input. This design leverages visual prompting as a lightweight yet effective way to enhance long-video understanding without introducing additional complexity.

\paragraph{Composite images.}
Combining multiple images into a single representation has been explored in several domains, such as multi-sensor fusion for autonomous driving~\citep{li2024llava}, efficient retrieval~\citep{nishimura2024vision}, and comic book understanding~\citep{iyyer2017amazing}. In the context of short video understanding, this strategy enables the usage of image models for video tasks, such as in action recognition~\citep{fan2021can} and question answering~\cite{kim2024image}.
These works demonstrate the effectiveness of spatially arranging frames to capture relationships across multiple instances. While we rely on the same principle, our work differs in several key aspects. First, we are the first to apply this idea to long video understanding. Second, we use video-language models, rather than image models, and demonstrate that composite images are efficient representation even for video-language models. Third, we are the first to fine-tune VLMs with these inputs, showing that it provides a better representation for long videos in general, rather than only in the zero-shot case.

\section{Method}

We focus on improving video understanding of VLMs as measured by their question answering accuracy. For this scenario, we make use of datasets in the form of $\{\x_i, q_i, y_i\}_{i=1}^N$ with videos $\x \in \mathbb{R}^{D\times 3\times H\times W}$ of duration D, multiple-choice questions $q \in \Sigma^*$ from an alphabet $\Sigma$, and correct answers $y$. The VLM is typically soft-constrained by a prompt to only output the letter of the correct answer, allowing for an easy evaluation and comparison between methods by measuring their accuracy~\citep{bai2025qwen2,fu2025video}. 

\subsection{Motivation}

Video understanding aims to determine the extent to which VLMs can reason about the videos they receive as input. While for short videos, this often comes in the form of action recognition, longer videos allow for higher-level questions. This makes them especially interesting, as reasoning aspects such as event ordering and long-range dependencies can now be considered. However, this increase in the sophistication of the questions comes with a notable increase in difficulty.

The main challenge is the limited temporal resolution of the VLMs, which is defined by their \textit{context window} $C$ and determines how many frames (or tokens) the model can process. In the case of long videos, i.e., the case where $D \gg C$, this means that the model can not densely parse the entire input video. Instead, the temporal resolution has to be reduced in order to fit within this constraint. In practice, this is done by a sampling function $\phi : \mathbb{R}^{D\times 3\times H\times W} \to \mathbb{R}^{T\times 3\times H\times W}$ that samples $T$ frames from the input video. 


The application of $\phi$ to long videos creates an important imbalance. VLMs are largely trained on images and short videos, where the spatial and temporal resolution are both still high after sampling. For long videos, however, the temporal resolution decreases drastically, while the spatial resolution remains the same. This mismatch leads to a disproportionately large part of the computational power of the model being dedicated to spatial rather than temporal relations.

\subsection{Visual Prompting for Long Video Understanding}

To re-balance the allocation of computational resources between the spatial and temporal components, we introduce the first visual prompting technique for long video understanding by combining multiple frames into a single image. 
By creating a sequence of these multi-panel frames, we enable denser temporal coverage within the given input budget, thereby increasing temporal resolution at the expense of some spatial detail. 
Beyond simply seeing more frames, this allows the model to use its strong pre-trained visual encoder to infer temporal relations as well. In this way, our approach effectively extends the video context of existing VLMs.
Our method operates in two steps:

\textbf{Dynamic frame sampling.}
The requirements for understanding short and long videos differ~\citep{li2024videochat}, and our prompting strategy is designed explicitly for long videos.
To reflect this, we choose the number of sampled frames $T$ dynamically depending on the ratio between the context window size $C$ and the video duration $D$:
\begin{equation}
\label{eq:fps}
T = \begin{cases*}
  C, & if $\gamma C \geq D$,\\
  \alpha \beta C,              & otherwise.
\end{cases*}
\end{equation}
As a result, we only use paneling in the case where the sampled frames are spaced at least $\gamma$ frames apart. 
The hyperparameters $\alpha$ and $\beta$ define the number of frames that are combined horizontally and vertically, respectively, into one image. We find that our method reaches the best results when $\alpha{=}\beta$, and, therefore, these can be considered as a single hyperparameter. Fig.~\ref{fig:method} shows our default setting with $\alpha{=}\beta{=}2$.
We explore other combinations of the hyperparameters in the ablation study.

\textbf{Panel construction.}
In the case where $\gamma C < D$, the resulting sampled video $\x \in \mathbb{R}^{\alpha \beta C\times 3\times H\times W}$ is too large for the context window of the VLM. Therefore, we downsample $\x$ to $\x' \in \mathbb{R}^{\alpha \beta C\times 3\times H / \alpha \times W / \beta}$. Then, every $\alpha \beta$ frames are stacked into one image in left-to-right, top-to-bottom order into a \textit{panel image}. This way, the final input to the VLM, $\x'' \in \mathbb{R}^{C\times 3\times H\times W}$, is again within the constraints expected by the vision encoder. For example, in the case of $\alpha{=}\beta{=}2$, we have:
\[
\x''_i =
\begin{pmatrix}
\x'_{4i} & \x'_{4i+1} \\
\x'_{4i+2} & \x'_{4i+3}
\end{pmatrix}.
\]
This procedure preserves the standard input shape while extending temporal coverage by a factor of $\alpha \beta$.

We found that for specific models, describing the panels in the prompt improved results. However, we did not find a single textual prompt that increases performance across models uniformly. As a result, we do not provide any extra information in the prompt. Nonetheless, if the model architecture is fixed, an appropriate description can improve results further. We explore this further in the supp.\ material.

\subsection{Fine-tuning}

Applying our approach to existing VLMs improves zero-shot performance on long-video benchmarks, even though these models are originally trained on standard videos without panels. 
When additional resources are available, VLMs can be fine-tuned to better adapt to the proposed input format and further enhance performance. 
Specifically, we fine-tune the models on the video-question pairs from their original training data by maximizing the likelihood of the correct multiple-choice answer:
\begin{equation}
    \ell_{FT}(\mathbf{x}, q, y) = -\log p_\theta(y \mid \mathbf{x}, q).
\end{equation}
We discuss the impact of fine-tuning in Section~\ref{sec:finetune}.


\section{Experiments}

We conduct experiments across five well-established datasets and eight models to evaluate our visual prompting strategy, demonstrating that it is model-agnostic and generally applicable. 

\subsection{Experimental Set-up}

\textbf{Models.}
Since our approach operates at the input level and does not require any architectural changes, we can apply it directly to a wide range of existing VLMs. In particular, we focus on three main groups of VLMs, which include the current state-of-the-art open-source models: 
\begin{itemize}
    \item \textbf{Small-context.} VLMs with up to 16 frames as input. We evaluate Video-LLaVA~\citep{lin2023video} and VideoChat2-HD~\citep{li2024mvbench}.
    \item \textbf{Medium-context.} 16-128 frames as input. We evaluate three variants of LLaVA-OV~\citep{li2024llava} (0.5, 7, and 72 billion parameters), Qwen-2.5VL~\citep{bai2025qwen2} with 32 input frames, and LLaVA-Video~\citep{zhangllava} in two variants (7 and 72 billion).
    \item \textbf{Long-context.} VLMs taking more than 128 frames as input. We evaluate Qwen-2VL~\citep{wang2024qwen2}, Qwen-2.5VL~\citep{bai2025qwen2}, and VideoLLaMA 3~\citep{zhang2024long}, all with 180 frames input.
\end{itemize}
Additionally, we report experiments on one of the datasets with two variants of the commercial ChatGPT model, with 8 and 32 frames as input.

\noindent\textbf{Datasets.}
We use five established and public benchmarks for our evaluation:
\textbf{VideoMME}~\citep{fu2025video} (VMME), \textbf{TimeScope}~\citep{timescope}, \textbf{MLVU}~\citep{zhou2025mlvu}, \textbf{MF2}~\citep{zaranis2025movie}, and \textbf{VNBench}~\citep{zhao2024needle}.
Extended details on the datasets are provided in the supp.\ material.
We report the accuracy for all questions across each benchmark. 

\noindent\textbf{Implementation details.}
We evaluate all models with \texttt{lmms-eval}~\citep{lmms_eval2024,zhang2024lmmsevalrealitycheckevaluation}. The VLMs are given the question and the multiple-choice answers in the format expected by the model, after which all VLMs get the standard instruction to ``Answer with the option's letter from the given choices directly.\textbackslash n''. 
We set $\alpha{=}\beta{=}2$, $\gamma$ to frames per second of the input video, and use uniform sampling for $\phi$. 
For fine-tuning, we optimize LLaVA-OneVision 7B on the video subset of LLaVA-Video-178K~\citep{zhangllava} for one epoch, using a batch size of 2 and a gradient accumulation of 4.   

\begin{table*}[t]
  \resizebox{\textwidth}{!}{
    \begin{tabular}{ccccccccccc}
    \toprule
           & \#frames & \multicolumn{3}{c}{\textsc{VMME}} & \multicolumn{2}{c}{\textsc{TimeScope}} & \textsc{MLVU} & \textsc{MF2} & VNBench & Avg. \\
\cmidrule(lr){3-5} \cmidrule(lr){6-7} & & medium & long & overall & Short & Long \\
 Avg. Duration (s) & & 516 & 2467 & 1018 & 2586 & 27600 & 651 & 5300 & 57 \\
\hline
\rowcolor{lgray} & \multicolumn{10}{c}{\textit{Small-context VLMs}} \\
Video-LLaVA 7B & 8 & 36.6 & 32.6 & 37.1 & 24.4 & \textbf{17.6} & \textbf{45.7} & \textbf{50.4} & 27.8 & 33.8 \\
\quad \textbf{+ ours} & 8 & \textbf{37.9} & \textbf{34.2} & \textbf{38.7} & \textbf{25.6} & 17.1 & \textbf{45.7} & 50.2 & \textbf{32.0} & \textbf{34.8} \textcolor{green!80!black}{(+1.0)} \\
\hline
VideoChat2-HD & 16 & 26.4 & 24.8 & 25.2 & 21.2 & \textbf{19.8} & 49.2 & \textbf{50.0} & 27.9 & 32.2 \\
\quad \textbf{+ ours} & 16 & \textbf{26.7} & \textbf{25.1} & \textbf{25.4} & \textbf{21.3} & \textbf{19.8} & \textbf{49.8} & \textbf{50.0} & \textbf{28.5} & \textbf{32.5} \textcolor{green!80!black}{(+0.3)} \\
\hline
\rowcolor{lgray} & \multicolumn{10}{c}{\textit{Medium-context VLMs}} \\
LLaVA-OV 0.5B & 32 & 40.9 & \textbf{37.0} & 43.8 & 49.4 & 25.6 & \textbf{44.0} & 50.1 & 39.8 & 42.1 \\
\quad \textbf{+ ours} & 32 & \textbf{42.6} & 36.6 & \textbf{44.3} & \textbf{56.9} & \textbf{30.0} & 43.1 & \textbf{50.2} & \textbf{41.0} & \textbf{44.3} \textcolor{green!80!black}{(+1.2)} \\
\hline
LLaVA-OV 7B & 32 & \textbf{56.7} & 48.8 & 58.5 & 58.7 & 30.2 & 62.9 & 51.5 & 54.8 & 52.8 \\
\quad \textbf{+ ours} & 32 & 56.2 & \textbf{50.2} & \textbf{58.9} & \textbf{69.5} & \textbf{33.8} & \textbf{65.3} & \textbf{52.1} & \textbf{57.7} & \textbf{56.2} \textcolor{green!80!black}{(+3.4)} \\
\hline
LLaVA-OV 72B & 32 & 62.9 & 57.6 & 66.0 & 59.1 & \textbf{33.8} & 21.6 & 56.6 & 59.4 & 49.4 \\
\quad \textbf{+ ours} & 32 & \textbf{66.4} & \textbf{59.3} & \textbf{67.7} & \textbf{70.0} & 32.4 & \textbf{23.8} & \textbf{58.5} & \textbf{62.6} & \textbf{52.5} \textcolor{green!80!black}{(+3.1)} \\
\hline
Qwen-2.5VL 7B & 32 & 61.6 & 51.2 & 61.9 & 52.8 & 28.7 & 60.1 & 52.6 & 55.6 & 51.9 \\
\quad \textbf{+ ours} & 32 & \textbf{64.0} & \textbf{54.7} & \textbf{63.9} & \textbf{60.8} & \textbf{30.0} & \textbf{64.9} & \textbf{53.8} & \textbf{58.5} & \textbf{55.3} \textcolor{green!80!black}{(+3.4)} \\
\hline
LLaVA-Video 7B & 64 & \textbf{62.3} & 53.6 & 64.3 & 64.8 & 34.7 & \textbf{66.2} & 52.8 & - & 56.6 \\
\quad \textbf{+ ours} & 64 & 62.2 & \textbf{54.0} & \textbf{64.4} & \textbf{79.2} & \textbf{39.3} & 66.0 & \textbf{54.4} & - & \textbf{60.7} \textcolor{green!80!black}{(+4.1)} \\
\hline
LLaVA-Video 72B & 64 & 67.8 & 61.2 & 69.8 & 65.4 & \textbf{30.9} & 52.9 & 58.2 & - & 55.4 \\
\quad \textbf{+ ours} & 64 & \textbf{68.6} & \textbf{61.4} & \textbf{70.1} & \textbf{75.7} & \textbf{30.9} & \textbf{54.4} & \textbf{59.9} & - & \textbf{58.2} \textcolor{green!80!black}{(+2.8)} \\

\hline
\rowcolor{lgray} & \multicolumn{10}{c}{\textit{Long-context VLMs}} \\
Qwen-2VL 7B & 180 & 62.7 & 51.0 & 62.4 & 66.1 & 23.8 & 65.7 & 54.3 & - & 54.5 \\
\quad \textbf{+ ours} & 180 & \textbf{62.9} & \textbf{52.7} & \textbf{63.0} & \textbf{71.2} & \textbf{26.7} & \textbf{65.8} & \textbf{56.7} & - & \textbf{56.7} \textcolor{green!80!black}{(+2.2)} \\
\hline
Qwen-2.5VL 7B & 180 & \textbf{67.6} & 54.8 & 66.0 & 73.9 & \textbf{37.6} & 66.7 & 54.2 & - & 59.7 \\
\quad \textbf{+ ours} & 180 & 66.9 & \textbf{56.1} & \textbf{66.3} & \textbf{79.1} & 35.6 & \textbf{66.8} & \textbf{54.8} & - & \textbf{60.5} \textcolor{green!80!black}{(+0.8)} \\
\hline
VideoLLaMA 3 7B & 180 & \textbf{64.6} & 54.1 & \textbf{65.3} & 80.2 & 39.1 & \textbf{47.3} & \textbf{58.9} & - & 58.2 \\
\quad \textbf{+ ours} & 180 & 63.7 & \textbf{55.1} & \textbf{65.3} & \textbf{87.2} & \textbf{46.7} & 47.1 & 58.3 & - & \textbf{60.9} \textcolor{green!80!black}{(+2.7)} \\
\hline
\rowcolor{lgray} & \multicolumn{10}{c}{\textit{Commercial VLMs}} \\
GPT-4o-mini-2025-04-14 & 8 & 50.0 & 45.6 & 51.1  & - & - & - & - & - & - \\
\quad \textbf{+ ours} & 8 & \textbf{53.0} & \textbf{46.9} & \textbf{53.6}  & - & - & - & - & - & - \\
\hline
GPT-4.1-2025-04-14 & 32 & 68.9 & 66.4 & 72.5  & - & - & - & - & - & - \\
\quad \textbf{+ ours} & 32 & \textbf{72.9} & \textbf{66.6} & \textbf{73.6} & - & - & - & - & - & - \\
\hline
    \end{tabular}
    }
     \caption{\textbf{Benefits of using our approach across datasets and models.} We report the video question answering accuracy. 
     Since the videos of VNBench are very short, we evaluate only VLMs with up to 32 frames of context. Our method improves the baseline results in nearly all cases and on average for all VLMs.
     }
  \label{tab:main_results}%
\end{table*}

\subsection{Zero-Shot Results}

We show in Table~\ref{tab:main_results} that our paneling approach consistently improves upon the baseline across a wide range of model architectures, parameter counts, context windows, and datasets, and using panels is, on average, the best approach for all models tested. The benefits are consistent across question types, but are especially pronounced in datasets where the task is to reason about ``needles", i.e., specific moments inserted in a video, such as TimeScope or VNBench. For example, paneling raises the VNBench score of LLaVA-OneVision 72B by $3.2$ percentage points, and the VMME and long TimeScope scores are raised by $2.0$ and $1.3$ points, respectively, for Qwen-2.5 VL with 32 context frames. For VideoLLaMA 3 7B, our approach even raises the accuracy from $39.1$ to $46.7$ on long TimeScope, which is an increase by $7.6$ points or $19.4\%$.     
In a few cases, our baselines achieve slightly higher results than previously reported, \eg, Qwen2.5-VL in the long-context setting on VideoMME~\citep{bai2025qwen2} ($65.1$ versus $66.0$). 
Even there, our panels improve the results further to $66.3$, showing that even with close to optimal settings, we can increase performance.
Nonetheless, there are still some cases where paneling slightly hurts the results, such as the performance of VideoLLaMA 3 7B on MF2 dropping by $0.6$ points, and some other isolated decreases in performance can be found. However, these cases do not occur often and are to be expected with such a wide evaluation. 

\begin{table*}[tb]
  \centering
    \begin{tabular}{lcccccccccc}
        \toprule
        & \multicolumn{4}{c}{\textsc{VMME}} & \multicolumn{2}{c}{\textsc{TimeScope}} \\
        \cmidrule(lr){2-5} \cmidrule(lr){6-7} & short & medium & long & overall & Short & Long \\
        \hline
        \rowcolor{lgray} \multicolumn{7}{c}{\textit{Without Panels}} \\
        No Fine-Tuning & 70.1 & 56.7 & 48.8 & 58.5 & 58.7 & 30.2 \\
        Proj + LLM & 71.1 & 56.2 & 48.2 & 58.5 & 58.0 & 30.9 \\
        \hline
        \rowcolor{lgray} \multicolumn{7}{c}{\textit{With Panels}} \\
        No Fine-Tuning & 70.1 & 56.2 & 50.2 & 58.9 & \textbf{69.5} & 33.8 \\ 
        Proj + LLM & 70.4 & 58.1 & 49.4 & \textbf{59.3} & \textbf{69.5} & \textbf{34.4} \\
        \hline
    \end{tabular}%
     \caption{\textbf{Effect of fine-tuning.} We report results on VMME using LLaVA-OV 7B. \textit{Proj} is the projector between the vision encoder and the Large-Language Model (LLM). Fine-tuning with panels reaches superior performance over fine-tuning with standard inputs.}
  \label{tab:finetune}%
\end{table*}

If we compare the averages over the datasets for each VLM, we observe a consistent improvement. For the VLMs with small context, \ie, $8$-$16$ frames, the average improvement is between $0.3$ and $1.0$ points. For the medium-context VLM, the average improvement is even higher and between $1.2$ and $4.1$ points. For the VLMs with $180$ frames as context, the gain is between $0.8$ and $2.7$ points. It is interesting to see that our approach with LLaVA-Video 7B, which uses 64 context frames, outperforms the long-context base VLMs. If we add our approach to the long-context VLMs, they remain the best performing approaches, but the gap is much smaller ($60.9$ for VideoLLaMA 3 7B vs.\ $60.7$ LLaVA-Video 7B). This shows on the one hand that our approach is versatile and improves various VLMs for video question answering without any additional training or increase of parameters. On the other hand, it questions whether existing VLMs efficiently utilize the larger context. 

We performed additional experiments with two closed-source models: GPT-4o-mini-2025-04-14 with 8 frames per video, and GPT-4.1-2025-04-14 with 32 frames per video. The results in Tab.~\ref{tab:main_results} confirm that our method also improves results for these commercial models on VMME by $2.5$ and $1.1$ points, respectively.

\subsection{Fine-Tuning Results}
\label{sec:finetune}

We show the results for fine-tuning with and without the proposed paneling in Table~\ref{tab:finetune} to disentangle the effect of paneling from the fine-tuning itself.
We find that fine-tuning further boosts performance by $0.4$ and $0.6$ on VMME and long TimeScope, respectively, over its zero-shot application when fine-tuning the projection and LLM modules. 
In the case without paneling, the average performance on VideoMME remains unchanged, while for Timescope, a slight improvement for long videos, but an equivalent drop for short videos, yields no net benefit. Overall, these results show that paneling images is an effective representation for long videos in general, both in the zero-shot setting and with fine-tuning.

\subsection{Panels vs.\ Pooling}
\label{sec:why}

Our approach offers a natural way to allocate more tokens to temporal rather than spatial resolution. While our approach is model-agnostic, as visual prompting is performed at the input level, there are other ways to achieve this. To further justify the choice of visual prompting with panels, we compare our method with another approach for increasing the ratio of tokens used for temporal understanding. 

We perform the analysis with LLaVA-OneVision and LLaVA-Video. 
We use average pooling on the visual tokens immediately before projecting them into the LLM. As there are $27\times27$ tokens per image at this point, we zero-pad to $28\times28$ before performing the pooling operation. This way, this baseline uses slightly more tokens than our proposed panels ($25088$ vs.\ $23328$).
We refer to this baseline in both cases as \textit{low-res}. 
For a fair comparison, we only apply \textit{low-res} when the videos are at least $\gamma C$ frames long (Eq.~\eqref{eq:fps}) and use normal sampling otherwise, similar to our own method.

\begin{figure}[tb]
    \centering
    \includegraphics[width=\linewidth, draft=false]{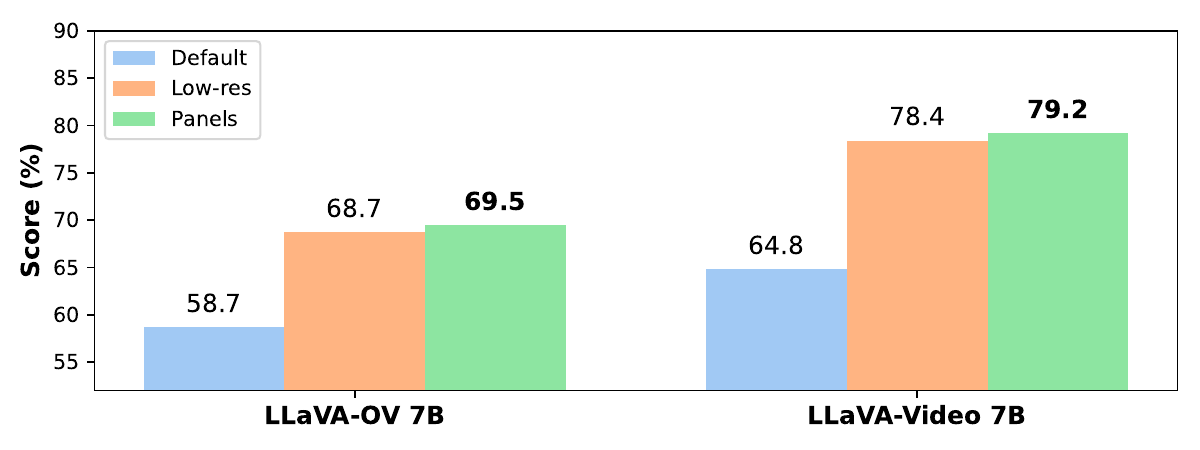}
    \vspace{-4mm}
    \caption{\textbf{Comparison to token pooling baseline.} We show results for LLaVA-OneVision 7B and LLaVA-Video 7B on TimeScope with three different strategies: default (no pooling), \textit{low-res}, and panels. Panels consistently improve the results.}
    \label{fig:just}
\end{figure}

We show the results of our comparison in Fig.~\ref{fig:just}. Our strategy outperforms both the default strategy and \textit{low-res} for all the models, except for LLaVA-Video 7B on TimeScope, where \textit{low-res} reaches equal performance. Overall, these results demonstrate that our approach is the optimal method for increasing temporal resolution, effectively allowing VLMs to utilize context sizes significantly larger than those with which they were trained. Additionally, we attempted to fine-tune models on these pooled representations. However, this led to quick overfitting and performance drops.

Our findings are in line with previous works on high-resolution image understanding with VLMs, which found that scaling resolution is more effective than scaling the number of tokens~\citep{li2024llava,liu2024llavanext}. Our approach can be seen as the inverse of this strategy; rather than splitting one high-resolution image into multiple chunks, we combine many high-resolution frames into one image, outperforming approaches based on token pooling.

\subsection{Ablations}
\label{sec:abl}

\begin{figure}[tb]
\vspace{-2mm}
  \centering
  \setlength\tabcolsep{10pt}
  \begin{tabular}{c}
    \includegraphics[width=0.95\linewidth]{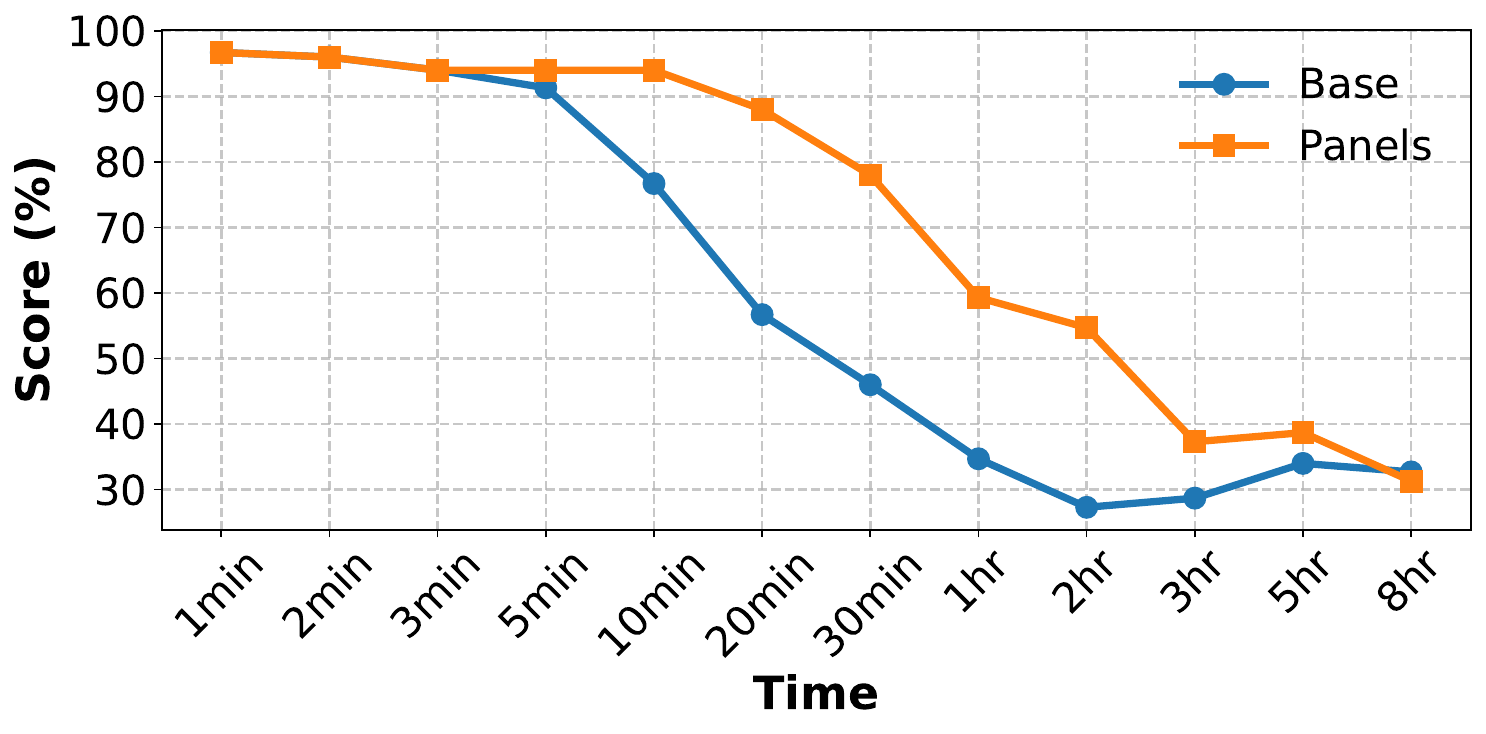} \\
    \includegraphics[width=0.95\linewidth]{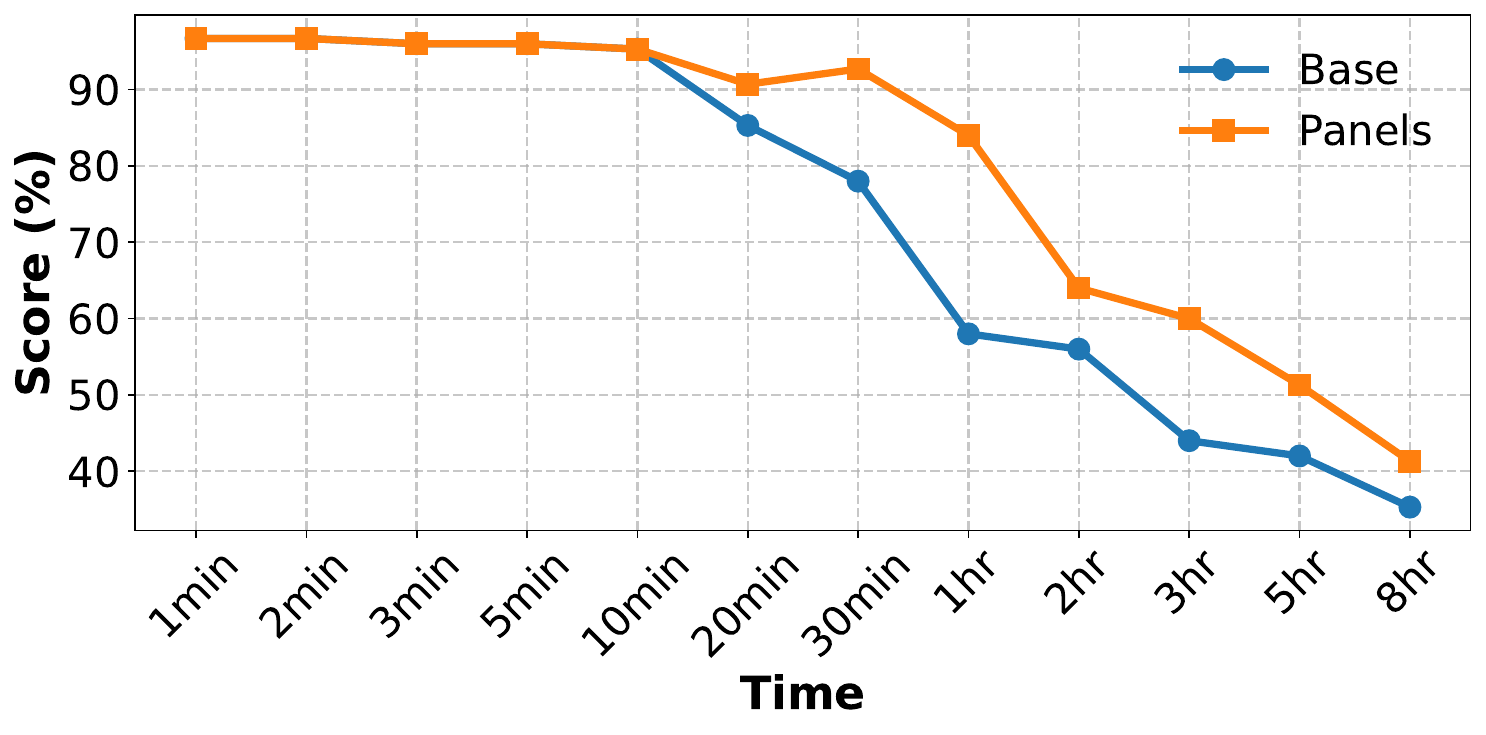} \\
  \end{tabular}
  \vspace{-2mm}
    \caption{\textbf{Performance as a function of video duration.} We show results for (top) LLaVA-Video 7B and (bottom) Video-LLaMA 7B on short and long TimeScope. Panels consistently improve the results for longer videos. }
    \label{fig:ts}
\end{figure}

\textbf{Performance for different video lengths.}
TimeScope~\citep{timescope} provides videos with corresponding question–answer pairs across $13$ different video lengths, allowing us to analyze how paneling affects performance as duration increases.
The results for LLaVA-Video 7B and Video-LLaMA 7B in~Fig.~\ref{fig:ts} show that our visual prompting strategy consistently improves accuracy on longer videos, confirming the efficacy of paneling inputs.


\begin{figure}[tb]
  \centering
  \begin{tabular}{c}
    \includegraphics[width=0.95\linewidth]{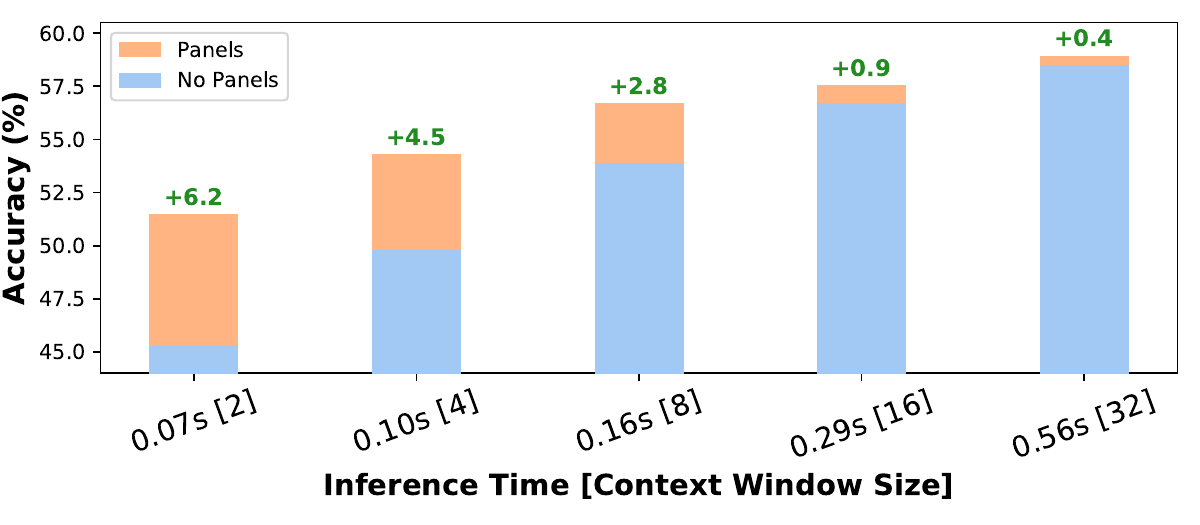} \\
    \includegraphics[width=0.95\linewidth]{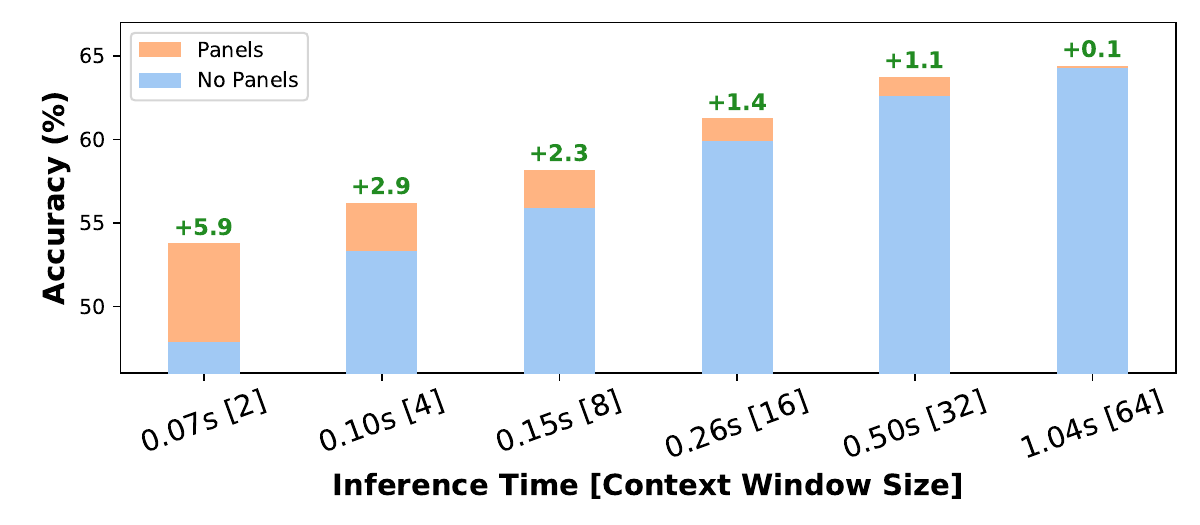} \\
  \end{tabular}
  \vspace{-2mm}
    \caption{\textbf{Performance as a function of context window size.} We show results for (top) LLaVA-OneVision 7B and (bottom) LLaVA-Video 7B. Panels consistently improve the results, with more pronounced effects for smaller window sizes.}
    \label{fig:context}
\end{figure}

\textbf{Trade-off between accuracy and inference time.} 
The same VLM can process different numbers of frames as input. Using fewer frames reduces the number of visual tokens to process, which in turn lowers inference time and computational cost. In Fig.~\ref{fig:context}, we show how paneling impacts performance across a range of different context window sizes, alongside the relative inference time. Our prompting strategy is effective for all window sizes. Furthermore, the gain in performance allows for fewer computational resources needed without losing performance. For instance, we achieve the same performance for LLaVA-OneVision 7B with half the frames ($8$ vs.\ $16$), and, therefore, also half the visual tokens.


\textbf{Impact of $\gamma$.}
We report results for different values of $\gamma$ in Tab.~\ref{tab:gamma}. Recall that lower values of $\gamma$ trigger paneling for shorter videos. 
Across all tested values, panels consistently outperform the baseline. However, skipping paneling for the shortest videos yields better results, as reflected in higher average performance for larger values of $\gamma$. 
Interestingly, from $\gamma=1\times$fps onwards, performance on short videos is slightly better than with no paneling at all, suggesting that paneling is already beneficial at relatively short videos.
We note that the impact of $\gamma$ is more pronounced on datasets with lower frame rates (\eg TimeScope at $2$ fps), where long durations are reached with fewer frames. 


\begin{table}[tb]
    \centering
    \begin{minipage}{0.48\textwidth}
        \centering
        \resizebox{\textwidth}{!}{
        \begin{tabular}{ccccccccccc}
            \toprule
            & \multicolumn{4}{c}{\textsc{VMME}} & \multicolumn{4}{c}{\textsc{TimeScope}} \\
            \cmidrule(lr){2-5} \cmidrule(lr){6-7} & short & medium & long & overall & Short & Long \\
            \hline
            No panels & 70.1 & 56.7 & 48.8 & 58.5 & 58.7 & 30.2 \\
            $\gamma=0$ & 69.2 & \textbf{56.8} & \textbf{49.7} & 58.6 & \textbf{70.5} & 33.8 \\
            $\gamma=0.5\times$fps &70.1 & \textbf{56.8} & \textbf{49.7} & \textbf{58.9} & 70.2 & 33.8 \\
            \rowcolor{blue!20!white} $\gamma=1\times$fps & \textbf{70.2} & \textbf{56.8} & \textbf{49.7} & \textbf{58.9} & 69.5 & 33.8 \\
            $\gamma=2\times$fps & \textbf{70.2} & \textbf{56.8} & \textbf{49.7} & \textbf{58.9} & 69.2 & 33.8 \\
            $\gamma=3\times$fps & \textbf{70.2} & 56.7 & \textbf{49.7} & \textbf{58.9} & 67.7 & 33.8 \\
            \hline
        \end{tabular}%
        }
        \caption{\textbf{Effect of the FPS constraint $\gamma$.} We report results on VMME using LLaVA-OV 7B. Results are better when not paneling on the shortest videos, but performance is robust to values of $\gamma$. The configuration used in our experiments is highlighted.}
        \label{tab:gamma}%
    \end{minipage}\hfill
    \begin{minipage}{0.48\textwidth}
        \centering
        \resizebox{\textwidth}{!}{
        \begin{tabular}{ccccccccccc}
            \toprule
            & \multicolumn{4}{c}{\textsc{VMME}} & \multicolumn{2}{c}{\textsc{TimeScope}} \\
            \cmidrule(lr){2-5} \cmidrule(lr){6-7} & short & medium & long & overall & Short & Long \\
            \hline
            1$\times$1 & 70.1 & 56.7 & 48.8 & 58.5 & 58.7 & 30.2 \\
            1$\times$2 & 69.9 & \textbf{57.6} & 48.4 & 58.6 & 65.9 & 31.3 \\
            2$\times$1 & \textbf{70.9} & 55.4 & 48.1 & 58.1 & 63.5 & 32.7 \\
            \rowcolor{blue!20!white} 2$\times$2 & 70.1 & 56.3 & \textbf{50.3} & \textbf{58.9} & 69.5 & \textbf{33.8} \\
            3$\times$3 & 70.1 & 56.1 & 49.9 & 58.7 & \textbf{76.5} & \textbf{33.8} \\
            4$\times$4 & 70.1 & 56.4 & 48.8 & 58.4 & 73.9 & 30.9 \\
            \hline
        \end{tabular}%
        }
        \caption{\textbf{Effect of the number of panels.} We report results on VMME using LLaVA-OV 7B. 2x2 panels achieve the best results, especially on long videos. The configuration used in our experiments is highlighted.}
        \label{tab:panels}%
    \end{minipage}
\end{table}

\begin{figure*}[tb]
    \centering
    \includegraphics[width=0.92\linewidth]{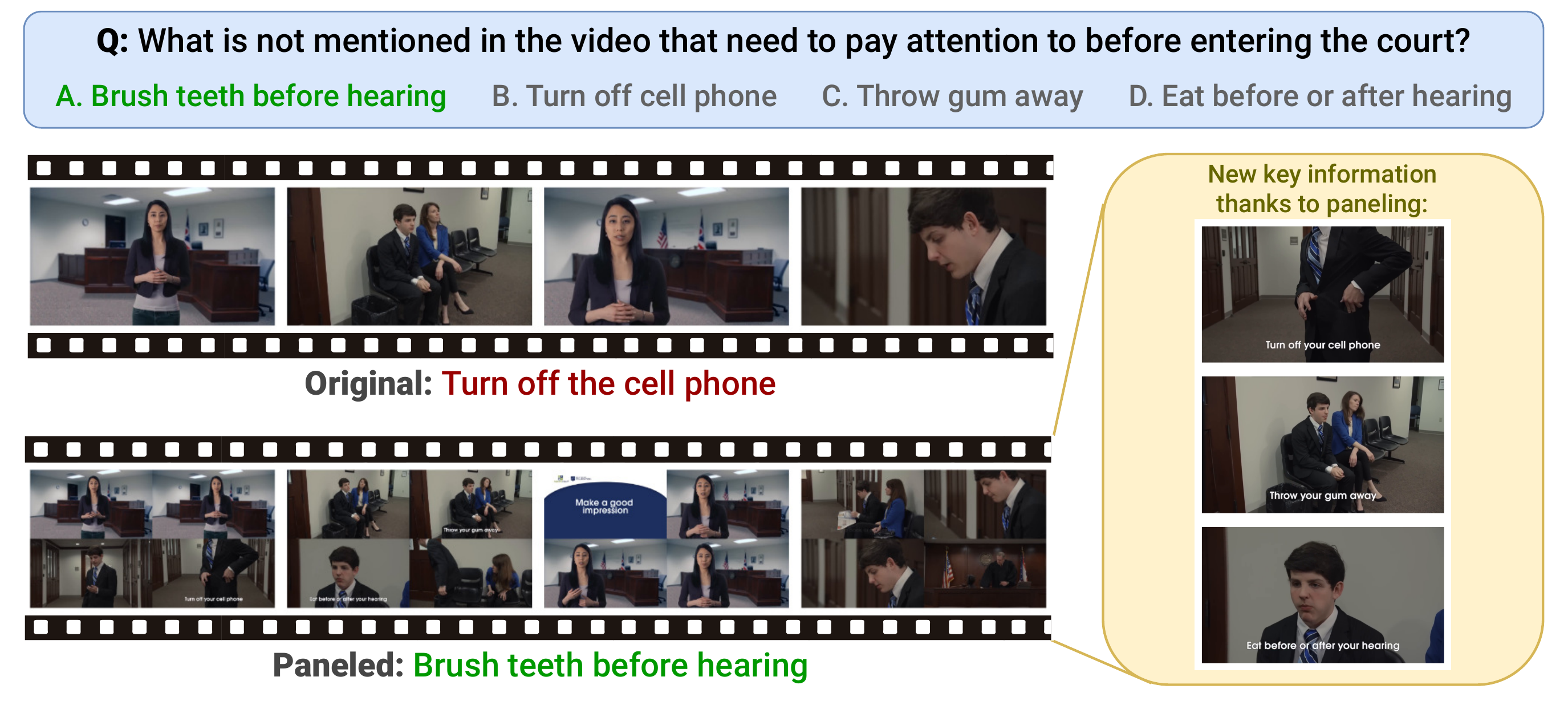}
    \caption{\textbf{Qualitative example on VideoMME.} We use LLaVA-OV 7B as the VLM. Without panels, the relevant information to answer the question is absent.}
    \label{fig:vmme-qual}
\end{figure*}

\textbf{Impact of $\alpha$ and $\beta$.}
We explore the effect of different values for $\alpha$ and $\beta$ on two benchmarks: VideoMME, which involves standard QA across various tasks, and TimeScope with needle-in-a-haystack (NIAH) questions. The results in Tab.~\ref{tab:panels} reveal several trends.  First, paneling itself improves performance overall in the majority of cases.
However, increasing both $\alpha$ and $\beta$ has a clear trade-off: it provides more frames to the VLM, which benefits benchmarks involving NIAH queries, but comes at the cost of reduced spatial resolution, which hurts tasks requiring fine-grained visual detail. 
Second, we find that uneven paneling, \ie, when $\alpha \neq \beta$, leads to worse results than when they are equal. 

\subsection{Native Long-video Models}

Some modern long video models, such as Qwen2.5~\cite{bai2025qwen2}, employ heuristics that dynamically select the number of frames and their token allocation based on video duration. Similarly, our paneling strategy in Eq.~\eqref{eq:fps} determines the number of frames given the video length. 
We show that our strategy can also improve performance for such a type of model: under the default settings of Qwen2.5 7B, where the upper limit of $768$ frames is consistently reached for the longer videos, the model achieves a performance of $65.1$ on Video-MME, as reported in the official paper~\cite{bai2025qwen2}. In contrast, our $180$-frame paneling setting reaches $66.3$, an improvement of $1.2$ points.

\subsection{Qualitative Results and Failure Analysis}
Paneling provides VLMs with a broader temporal context, allowing them to access more information. In the example of Fig.~\ref{fig:vmme-qual}, the key evidence for answering the question lies in the text appearing within the video frames. Despite the lower spatial resolution of each paneled frame, the VLM successfully identifies and interprets this information, demonstrating that enhancing temporal coverage at the cost of some spatial detail is beneficial for long-video understanding.

We analyze in what cases paneling hurts performance using the Qwen2.5 7B model with 32-frame input on the MLVU dataset, which categorizes its questions into seven types. We show the full results in the supp.\ material.
Panels outperform the standard representation in tasks involving needle, counting, egocentric videos, topic reasoning, and plot question answering. Interestingly, panels improve the results for counting from $23.3\%$ to $39.8\%$ accuracy, showing that, while panels sacrifice some spatial resolution, they are still able to improve in tasks requiring spatial detail. Both representations perform equally in anomaly recognition, and the standard representation performs slightly better than panels in the ordering task by $1.2\%$. While panels preserve the temporal ordering, the temporal information is not explicitly encoded, which is a limitation. We explore two ways to remedy this based on NumPro~\cite{wu2025number} in the supp.\ material.



\section{Conclusion}

We presented the first visual prompting strategy for long-video understanding. By combining multiple frames into panels within a single image, our method enhances the temporal resolution of existing VLMs. This training-free, model-agnostic approach can be integrated seamlessly without modifying the underlying architecture. 
Extensive experiments demonstrate that it consistently improves performance across a wide range of benchmarks and model designs.
Furthermore, we demonstrated that fine-tuning can yield additional gains, both compared to zero-shot paneling performance and fine-tuning with the standard data representation. While our approach does not improve the understanding capabilities of the underlying VLMs, it raises the bar for new models in long video understanding to surpass before being able to justify their additional complexity. 

\section*{Acknowledgments}
The work has been supported by the ERC Consolidator Grant FORHUE (101044724) and the Federal Ministry of Research, Technology and Space (BMFTR) under grant no. 01IS22094A WEST-AI. For the computations involved in this research, we acknowledge EuroHPC Joint Undertaking for awarding us access to Leonardo at CINECA, Italy, through EuroHPC Regular Access Call - proposal No.\ EHPC-REG-2025R01-218.

{
    \small
    \bibliographystyle{ieeenat_fullname}
    \bibliography{main}
}

\clearpage
\setcounter{page}{1}
\maketitlesupplementary



\section{Dataset Details}
\label{sec:data}

\begin{itemize}
    \item \textbf{VideoMME}~\citep{fu2025video} (VMME). A multi-modal evaluation benchmark that spans across $6$ different domains (Knowledge, Film \& Television, Sports Competition, Artistic Performance, Life Record, and Multilingual). Video durations range between $11$ seconds to $1$ hour, and are categorized as \textit{short} videos ($80$ seconds on average), \textit{medium} ($500$ seconds) and \textit{long} ($2500$ seconds). 
    \item \textbf{TimeScope}~\citep{timescope} inserts multiple short video clips (``needles'') into long videos. These needles contain the key information required to answer the questions. As opposed to the usual Needle In A Haystack (NIAH) evaluation, this forces models to understand the whole video. Videos are divided into $13$ different lengths, ranging from $60$ to $36,000$ seconds. We divide the evaluation into \texttt{short} (up to $3$ hours, average $2590$ seconds) and \texttt{long} (between $5$ and $10$ hours, average $27,600$).    
    \item \textbf{MLVU}~\citep{zhou2025mlvu} covers a wide range of video lengths, from $3$ minutes to $2$ hours, with an average length of $15$ minutes. It includes both real-world videos (\eg, egocentric, movies) and simulated videos (\eg, video games, cartoons). The benchmark provides multiple-choice and open-ended QA tasks across $9$ different categories. We evaluate on the \textit{dev} set, which consists of $2593$ QA pairs over $1730$ videos.
    \item \textbf{MF2}~\citep{zaranis2025movie} evaluates comprehension and recall of key narrative information from full-length movies ($50$-$170$ minutes). It includes $53$ complete movies with an average duration of $88.3$ minutes. Unlike other benchmarks, the task is to discriminate between true and false claims across $850$ claim-pairs. The claims span five categories: character motivations and emotions, memorable moments, causal chains, and event order.
    \item \textbf{VNBench}~\citep{zhao2024needle} (VNB) targets three aspects of video understanding: temporal perception, chronological ordering, and spatio-temporal coherence. It includes tasks such as retrieval, ordering, and counting, and consists of $1350$ samples ranging from $10$ to $180$ seconds, collected from $150$ videos. 
\end{itemize}

\section{Full Results for Failure Analysis}

\begin{table*}[tbp]
    \centering
    \resizebox{\textwidth}{!}{
    \begin{tabular}{ccccccccc}
        \toprule
         & Needle & Count & Ego & Topic Reas. & PlotQA & Anomaly Recogn. & Order \\
        \hline
Qwen2.5-7B & 71.0 & 23.3 & 55.7 & 85.6 & 63.1 & \textbf{72.0} & \textbf{49.8} \\
\quad \textbf{+ ours} & \textbf{72.1} & \textbf{39.8} & \textbf{56.8} & \textbf{89.0} & \textbf{65.3} & \textbf{72.0} & 48.6 \\
        \hline
    \end{tabular}%
    }
    \caption{\textbf{Analyzing failure cases on MLVU.} We report results for Qwen2.5 7B with 32 frames as input. 
    }
    \label{tab:mlvu}%
\end{table*}

We present the full results that have been discussed in Section 4.7, using the Qwen2.5 7B model with 32-frame input on the MLVU dataset in Tab.~\ref{tab:mlvu}.
As discussed in the main paper, panels outperform the standard representation in all tasks, except anomaly recognition, where they perform equally, and ordering, where performance drops by $1.2\%$. This highlights a limitation in the zero-shot panels representation, as temporal information is not explicitly encoded.
In the tasks where panels improve the results, the gain is typically around $1$-$4\%$. For the counting task, the performance increases from $23.3\%$ to $39.8\%$. This demonstrates that panels are also effective in tasks requiring spatial detail.

\section{Adding Explicit Temporal Encoding}

We conduct an experiment to explore whether incorporating some form of temporal encoding to video panels increases overall performance. 
We implement a temporal order encoding in two ways based on NumPro~\cite{wu2025number}, by overlaying the frame index (panels-F) and the frame time (panels-T) on the panels.
We evaluate these methods against the base video panels without explicit temporal encoding in Tab.~\ref{tab:temp}. The results show that adding explicit temporal information improves performance on temporal tasks such as answering questions about order. Adding the frame index (panels-F), however, reduces performance for some other types, such as counting. While adding the frame index to the panels overall decreases the performance, adding the frame time slightly increases the performance on MLVU. 
Thus, explicitly including temporal order is beneficial for specific question types and can be used when required by a specific task.

\begin{table}[htpb]
    \centering
    \begin{tabular}{cccc|c}
        \toprule
         & Count & PlotQA & Order & Overall \\
        \hline
        Qwen2.5-7B & 23.3 & 63.1 & 49.8 & 60.1 \\
        \quad \textbf{+ panels} & 39.8 & 65.3 & 48.6 & 63.4 \\
        \quad \textbf{+ panels-F} & 35.4 & 64.9 & {50.6} & 62.4\\
        \quad \textbf{+ panels-T} & 41.3 & 65.1 & {50.6} & {63.9} \\
        \hline
    \end{tabular}%
    \caption{\textbf{Effect of adding explicit temporal information.} We report results for Qwen2.5 7B with 32 frames on MLVU. 
    }
    \label{tab:temp}%
\end{table}

\section{Prompt} 
Our main results show that existing VLMs are already capable of interpreting the paneled images without any additional information in the prompt. Nonetheless, adding additional directions to the textual prompt can, in some cases, further improve results. 
We show results for LLaVA-OV 7B and Qwen2.5-VL 7B with three different prompts:
\begin{itemize}
    \item \textbf{Prompt 1} (P1): ``You are given a sequence of images. Each image is a composite grid of video frames arranged in temporal order: panels are ordered from left to right, then top to bottom — like reading a book. Within each composite, the panels represent consecutive frames from the video. Across the sequence, the composites are shown in chronological order. When answering, interpret the full temporal sequence, not individual panels in isolation." The prompt is added before the question.
    \item \textbf{Prompt 2} (P2): ``When answering, treat the panels as frames from one video, in order from left to right, then top to bottom." The prompt is added before the question.
    \item \textbf{Prompt 3} (P3): ``Each image is divided into \{r\} rows and \{c\} columns of panels. Read them in left-to-right top-to-bottom order as consecutive video frames. Answer with the option's letter from the given choices directly." The prompt is added after the question. \{r\} and \{c\} are replaced by the number of rows and columns.
\end{itemize}
As can be seen from Tab.~\ref{tab:prompts}, even among these models, there is no consistently best prompt. However, with model-specific prompts, performance can get another boost, such as Prompt 1 for VMME with LLava-OneVision, and Prompt 3 for VMME with Qwen2.5-VL. While model-specific prompts improve the results further, we do not include them for the results in the paper since our focus is on a model-agnostic approach.  

\begin{table}[t]
  \centering
    \begin{tabular}{cccccc}
        \toprule
         & No prompt & P1 & P2 & P3\\
        \hline        
        LLaVA-OV 7B & 58.9 & 60.1 & 59.4 & 58.8 \\
        Qwen2.5-VL & 62.4 & 61.9 & 61.8 & 62.9\\
        \hline
    \end{tabular}%
     \caption{\textbf{Effect of changing the prompt.} We report results of using additional prompts on VMME using LLaVA-OV 7B and Qwen2.5-VL. }
  \label{tab:prompts}%
\end{table}

\section{Qualitative Results}
\begin{figure*}[tbp]
    \centering
    \includegraphics[width=\linewidth]{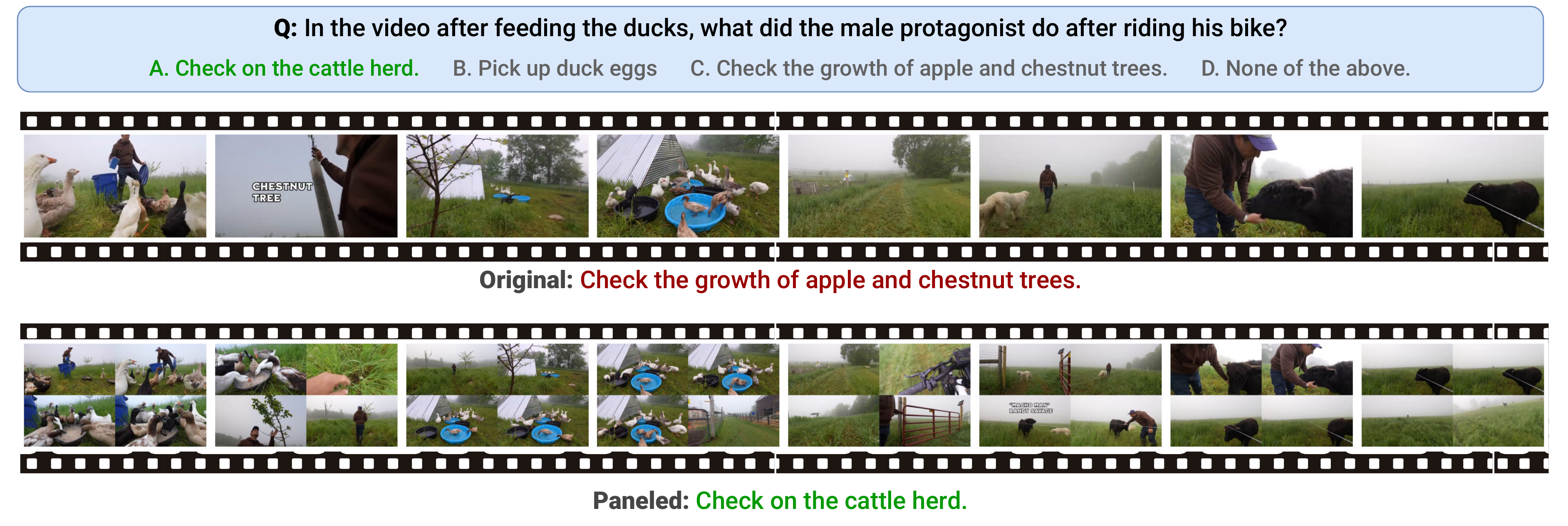}
    \caption{\textbf{Qualitative example on VideoMME.} We use LLaVA-Video 7B as the VLM. Without panels, the relevant information to answer the question (frames with the bike) is absent.}
    \label{fig:mme-qual-vid}
\end{figure*}

\begin{figure*}[tbp]
    \centering
    \includegraphics[width=0.8\linewidth]{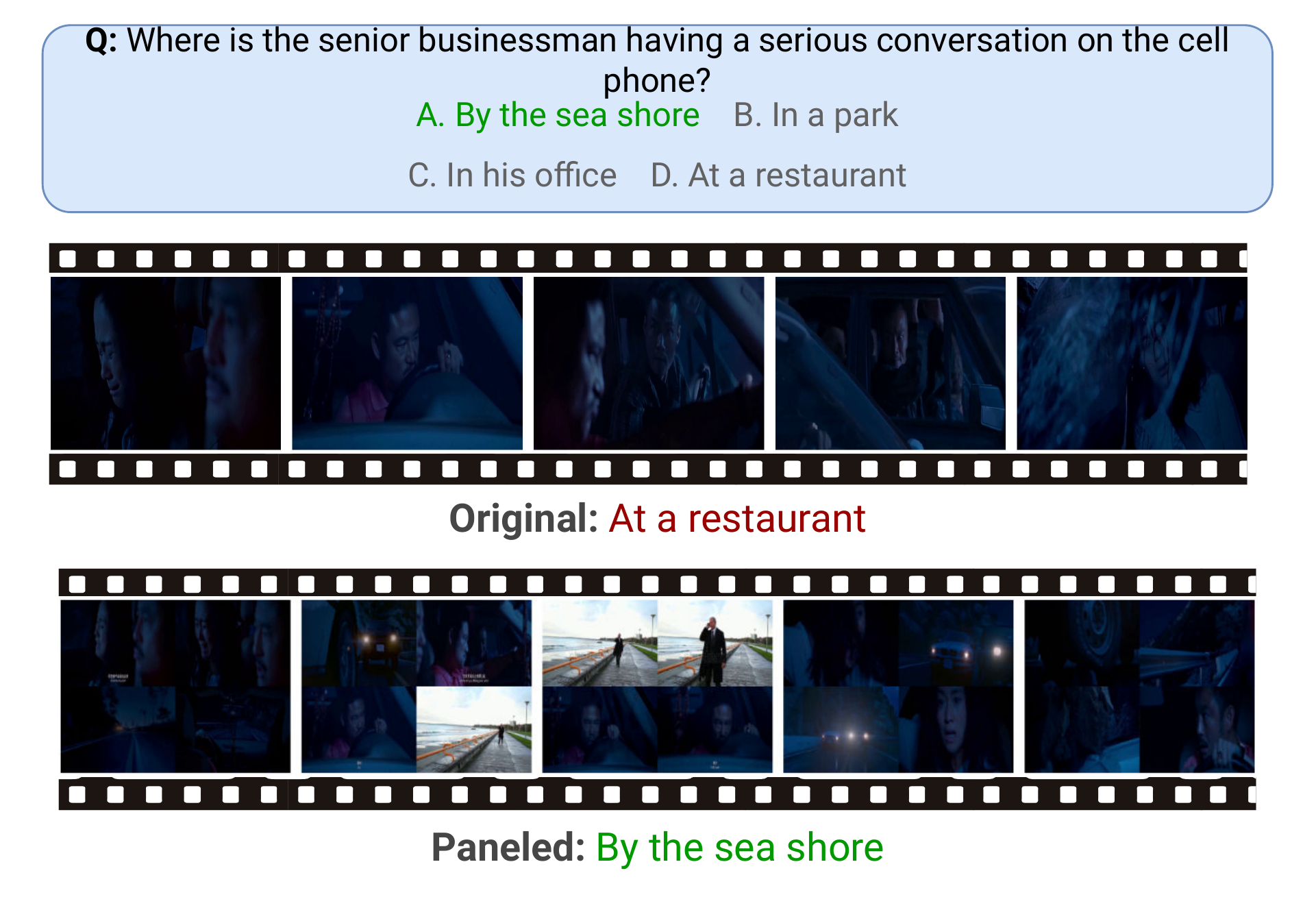}
    \caption{\textbf{Qualitative example on MLVU.} We use LLaVA-OV 7B as the VLM for the Needle In A Haystack task. Without panels, the relevant `needle' to answer the question is absent.}
    \label{fig:mlvu-qual-ov}
\end{figure*}

Figure~\ref{fig:mme-qual-vid} shows an example from the VideoMME benchmark using LLaVA-Video 7B as the VLM. The question asks what the protagonist does after \textit{feeding the ducks} and \textit{riding the bike}. Without our paneling strategy, the model processes only a sparse set of frames and observes only the `feeding the ducks' portion, entirely missing the segment where the person rides the bike. In contrast, our Video Panels capture both events by providing finer temporal coverage, allowing the model to correctly answer the question.
We also present an example of the challenging Needle In A Haystack (NIAH) task in Figure~\ref{fig:mlvu-qual-ov}, where the goal is to answer the question based on a very short, discriminative clip embedded within a much longer video. In this example, the question concerns a man walking along the shore. Without our paneling strategy, the crucial frames containing the target event are never passed to the VLM, preventing it from retrieving the correct evidence. In contrast, Video Panels ensures that these sparse but essential frames are included, enabling an accurate answer. 

\end{document}